\documentclass[letterpaper, 10 pt, conference]{ieeeconf}  

\IEEEoverridecommandlockouts                              
\usepackage[utf8]{inputenc}
\usepackage[T1]{fontenc}

\usepackage{graphicx} 
\usepackage{epsfig} 
\usepackage{mathptmx} 
\usepackage{mathptmx} 
\usepackage{amsmath} 
\usepackage{amssymb}  
\usepackage[dvipsnames]{xcolor}
\usepackage{float}
\usepackage[textsize=footnotesize]{todonotes}
\usepackage{multirow}
\usepackage[caption=false,font=footnotesize]{subfig}
\usepackage{xcolor}
\usepackage{array}

\usepackage{booktabs}
\usepackage{siunitx}
\usepackage{lipsum}
\usepackage{array}
\usepackage{gensymb}
\usepackage{cite}
\usepackage{hyperref}

\title{\LARGE \bf
A Novel Variable Stiffness Soft Robotic Gripper}

\author{Dimuthu~D.~Arachchige$^{1}$, Yue Chen$^{2}$, Ian D. Walker$^{3}$, and  Isuru~S.~Godage$^{1}$
\thanks{$\!\!\!\!\!\!\!\!\!\!^{1}$School of Computing, DePaul University, Chicago, IL 60604.\,{\tt\small DARACHCH@depaul.edu}\,
$^{2}$Dept. of Mechanical Engineering, University of Arkansas, Fayetteville, AR 72701.\,
$^{3}$Dept. of Electrical and Computer Engineering, Clemson University, Clemson, SC 29634.
\vspace{2mm}
\newline 
This work is supported in part by the National Science Foundation Grants IIS-1718755 and IIS-2008797.
\vspace{2mm}
\newline
This paper has been submitted to IEEE International Conference on Robotics and Automation 2021.
}
}

\begin{document}
\maketitle
\thispagestyle{empty}
\pagestyle{empty}

\begin{abstract}
We propose a novel tri-fingered soft robotic gripper with decoupled stiffness and shape control capability for performing adaptive grasping with minimum system complexity. The proposed soft fingers adaptively conform to object shapes facilitating the handling of objects of different types, shapes, and sizes. Each soft gripper finger has an inextensible articulable backbone and is actuated by pneumatic muscles. We derive a kinematic model of the gripper and use an empirical approach to map input pressures to stiffness and bending deformation of fingers. We use these mappings to achieve decoupled stiffness and shape control. We conduct tests to quantify the ability to hold objects as the gripper changes orientation, the ability to maintain the grasping status as the gripper moves, and the amount of force required to release the object from the gripped fingers, respectively. The results validate the proposed gripper's performance and show how stiffness control can improve the grasping quality.
\end{abstract}

\section{Introduction\label{sec:Introduction}}
Soft robots have higher flexibility and compliance than traditional rigid robots  \cite{chen2020modal, galloway2019fiber,shintake2018soft,walker2020soft}. Research shows that soft robotic grippers can handle objects of various shapes and sizes and different object categories that are predominantly fragile and highly deformable \cite{shintake2018soft,walker2020soft, manti2015bioinspired,zhou2017soft,chenl2018manipulation}. Therefore soft robotic grippers present a promising path toward adaptive grasping, which can mimic the dexterity and versatility of biological appendages such as human hands. With rigid grippers, it is challenging to know the contact forces without added sensors, and therefore force control may be needed to prevent damages to objects. In addition, stiffness or compliance control of rigid grippers requires complex actuator-elastic element arrangement. 

\begin{figure}[t] 
	\centering
	\includegraphics[width=1\linewidth]{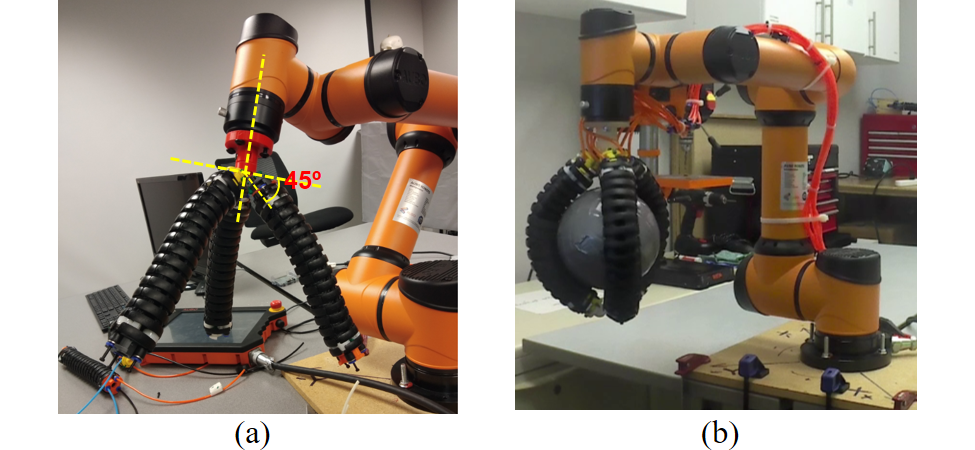}
	\caption{(a) Gripper attached as an end-effector of AUBO-i5 rigid robot, (b) Gripper grasping a sphere object.}
	\label{IntroductionImage} 
\end{figure}

In contrast, soft robotic grippers can conform to the environment to distribute contact forces (than large point forces) and, therefore, facilitate inherent compliant grasping without complex sensory or actuator arrangements. Further, finger conformity improves the grip quality while stiffness control, if available, helps lock and maintain the finger deformation. The gripper may need to exert large forces on objects depending on the type, shape, and weight. Here, stiffness control enables it to withstand a range of external loads while preserving the grip. Also, it helps retain the payload without compromising dexterity in manipulation. In this work, we propose a novel pneumatically actuated soft robotic gripper [Fig. \ref{IntroductionImage}] with adaptive and decoupled stiffness and shape control. The soft robotic fingers utilize an articulable rigid backbone to provide structural integrity, enable stiffness control while presenting a soft environmental interface. The finger actuators are assembled in a single unit and bring this additional functionality while simplifying the grasping operation \cite{hao2018design}. 

\subsection{Previous Work\label{subsec:Previouswork}}
Many soft robotic grippers have been proposed to-date. The work reported in \cite{liu2020novel} employs shape-memory alloys (SMA) and variable fluid viscosity to achieve variable stiffness. Similarly, authors of \cite{shintake2015variable} and \cite{yan2018design} used dielectric elastomer and low-melting-point alloy to assemble a variable stiffness gripper. Yang et al. \cite{yang2017novel} present a variable stiffness gripper with a built-in position feedback system. Their actuation principle and used materials are identical to the work of \cite{liu2020novel}. Thermally activated materials show a relatively slow phase transition. Therefore, the performance of these grippers in stiffness control is weak. Additionally, dielectric elastomers require a high operating voltage and have low efficiency due to the need for heating and cooling and associated energy loss. Chen et al. \cite{chen2020variable} present a layer-jamming induced active vacuum adhesion gripper that can lift flat, concave, and convex-shaped objects. Here, a particle pack acts as a stiffness-changeable gripper that is susceptible to membrane damages. The 3D-printed soft gripper in \cite{mutlu20173d} uses a combination of negative and positive pressures to vary the stiffness. The gripper reported in \cite{wang2018soft} inflates a portion of the soft finger to achieve stiffness control. Due to monolithic design, their actual force output makes them suitable for small-scale applications. In \cite{sun2020soft}, a pangolin scales inspired layer with toothed pneumatic actuators is employed to control stiffness. The gripper design of \cite{al2017design} uses antagonistically arranged pneumatic muscle actuators (PMAs) to achieve decoupled shape and stiffness variation. Further, the research in \cite{al2017design} uses a tendon-driven mechanism to apply forces to the fingers, which cause flexion and extension of the fingers. In \cite{tang2019development}, the gripper utilizes a screw and rod mechanism to complete the grasping operation, while PMAs control the stiffness. These mechanically driven grippers exhibit compliant grasping but have complex control schemes with low efficiency. We propose a novel variable stiffness gripper to address these challenges. 

\subsection{Contribution\label{subsec:Contribution}} 
This work's main contribution is to design and validate a novel variable stiffness soft gripper with decoupled shape and stiffness control with minimum system complexity. The gripper fingers incorporate an articulable and inextensible backbone with PMAs to achieve higher stiffness ranges that could be used to strong grips without trading compliance. The backbone is inspired by the highly adaptable biological tails of the spider monkey. The tail can act as a manipulator or supporting structure while standing upright and a balancing appendage during jumping and climbing. The tail's muscular lining actuates the skeletal structure underneath to obtain the desired structural stiffness and strength at these transformations. Similar to their biological counterparts, when PMAs of the proposed gripper fingers are actuated, the backbone offers structural integrity. This technique is novel for soft robotic grippers. We validate the gripper operation using three tests to assess the gripper efficacy, which shows that the proposed gripper can successfully grasp and maintain grips under various dynamic conditions.

\begin{figure}[tb] 
	\centering
	\includegraphics[width=8.6cm]{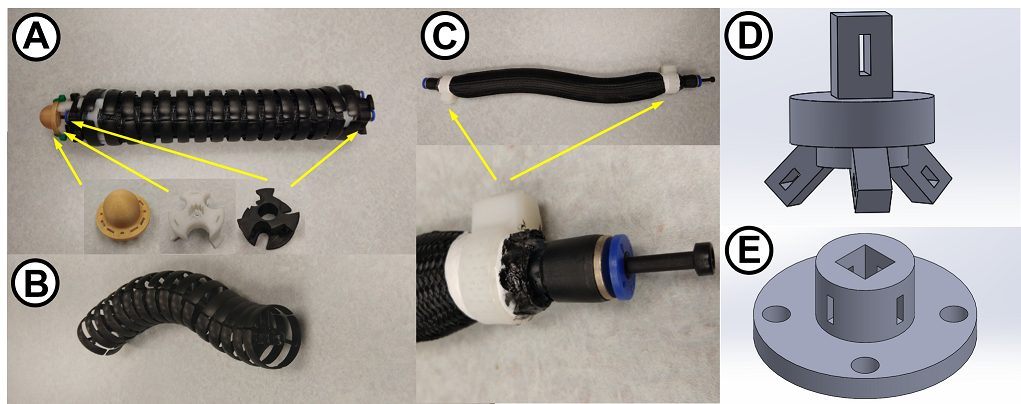}
	\caption{(A) Assembled soft robotic finger, (B) Inextensible backbone, (C) A PMA, (D)-(E) CAD models of gripper base unit used to assemble fingers. The top unit (D) connects the gripper to AUBO-i5 6-DoF industrial robot.}
	\label{CADPMAFingerGripper} 
\end{figure}

\section{Prototype Description\label{sec:prototypedescription}}
\subsection{Gripper Design and Fabrication\label{subsec:finger-design-fabrication}}
\subsubsection{Finger Design\label{subsubsec:finger-design}}
The completed soft finger is shown in Fig. \ref{CADPMAFingerGripper}-A. It has two main components; an articulable inextensible backbone (Fig. \ref{CADPMAFingerGripper}-B) and PMAs (Fig. \ref{CADPMAFingerGripper}-C). The backbone is a readily available robot dress pack manufactured by Igus Inc (part \# Triflex R-TRL40). In the design, individual high tensile strength plastic units have been assembled to form an energy chain. The individual units connected with ball-and-socket links allow free movement in multiple directions. We can 3D-print the same units for our purpose. However, it is rather cumbersome to 3D-print this complex unit as an entire backbone. Thus we decided to use the one available off the shell. We decided on the backbone's length based on its bending capacity to form a 180\textdegree subtended angle. PMAs are anchored to the backbone using 3D-printed anchor units shown in Fig. \ref{CADPMAFingerGripper}-A. Extension mode PMAs are custom made and 150~mm long. They can sustain pressures up to 700~kPa and extend by 50\%. Fabrication steps of PMAs are presented in \cite{godage2012pneumatic}. The inner diameter and thickness of the silicone PMA bladder tube are 11~mm and 2~mm, respectively.  The backbone is lined with three mechanically identical PMAs, push-to-connect fittings, 3D-printed finger end caps, and the wrapping of the whole unit (Fig. \ref{CADPMAFingerGripper}-A). A finger module has an effective length of 180~mm and a diameter of 36~mm. The backbone constraints the axial extension, and therefore a finger can only bend. The uniform arrangement of PMAs ensures circular arc bending within normal operating conditions. The backbone structure with PMAs helps to achieve independent stiffness control through antagonistic muscle arrangement.

\subsubsection{Gripper Fabrication\label{subsubsec:Gripper-fabrication}}
Our proposed gripper (Fig. \ref{IntroductionImage}) consists of three soft fingers in a symmetrical tri-fingered configuration to offer significant flexibility and versatility in handling different shapes of objects. The fingers are attached to a 3D-printed base unit shown in Fig. \ref{CADPMAFingerGripper}-D and E. The anchoring elements are designed at a $\frac{pi}{4}$~rad angle with respect to the horizontal plane (Figs. \ref{IntroductionImage}-a and \ref{schematicdiagram}-b). In fingers, we actuate two PMAs of the three simultaneously, which results in two effective DoF per finger. The reason for this decision is threefold; (1) obtain bidirectional bending using extension mode PMAs in an antagonistic arrangement,  (2) strong gripper force generation with simultaneous actuation of 2 PMAs to generate twice as much bending torque, and (3) achieve stiffness control with the remaining PMA. Therefore, this gripper design allows us to generate more gripping torque while decoupling stiffness and shape control. In the proposed gripper design, we simultaneously actuate all three fingers, and thus the gripper has two DoF in total. 

\section{System Modeling\label{sec:Modelling}}
\subsection{Kinematic Model\label{sub:kinematic-model}}

\begin{figure}[tb] 
	\centering
	\includegraphics{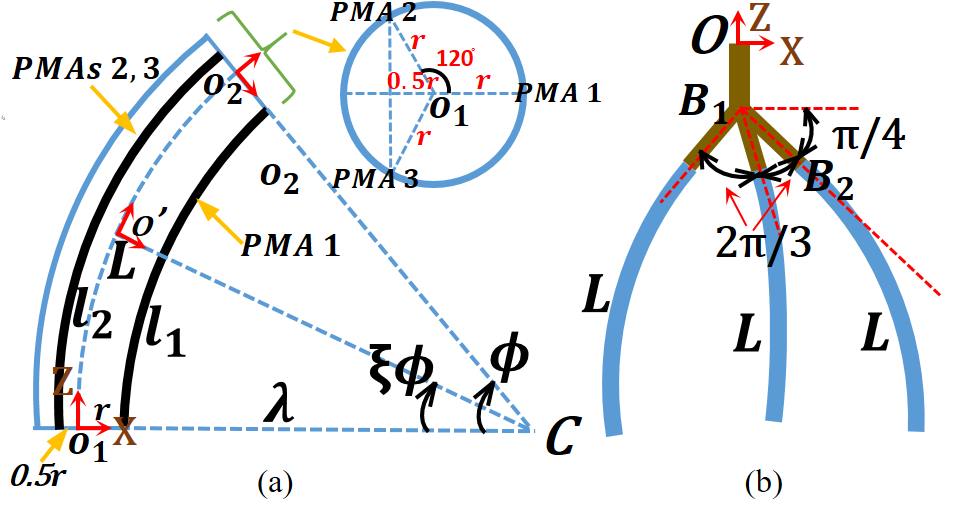}
	\caption{(a) Schematic of the soft robot arm, (b) Schematic of the gripper.}
	\label{schematicdiagram} 
\end{figure}

Fig. \ref{schematicdiagram}-a shows the schematic of the finger module. The finger base is positioned at the origin of the robot coordinate frame such that one PMA is on the X~axis and the length change of this is $l_1$. Thus, upon actuation, the finger bends on the X-Z~plane. Noting prior work, we assume that the finger module bends in a circular arc shape \cite{nazari2019forward,deng2019near}. The three PMAs are placed on a circular rigid frame at a radius $r$ from the center and $\frac{2\pi}{3}$ rads apart inside the inextensible rigid backbone (Fig. \ref{schematicdiagram}-a). Thus, actuation points of PMAs form an equilateral triangle of side $r\sqrt{3}$. With that, when one PMA (PMA 1) forms a $r$ distance along the X~axis from the  backbone center, the remaining two PMAs (PMAs 2 \& 3) form $\frac{r}{2}$ distance along the -X axis from the backbone center. $L_{0}$ denotes the initial length of the PMAs and, $l_{i}\in\mathbb{R}$ is the length variation where $l_{i:min}\leq l_{i}(t)\leq l_{i:max}$ for $i\in \left \{ 1,\ 2\right \}$ and $t$ denotes the actuator number and time, respectively. Thus, $L_{i}(t)=L_{0}+l_{i}(t)$ presents the length of each actuator at time $t$. The work reported in \cite{godage2011shape} presented curve parameters that define the spatial position and orientation of a single section extensible continuum arm in 3D taskspace. We can extend the derivation of \cite{godage2015modal} to obtain model kinematics of an inextensible finger in planar taskspace. Let the joint variable vector be ${q}=\left [ l_{1}(t),\ l_{2}(t) \right ]^{T}$. Two spatial parameters can describe the finger bending arc, the radius of curvature, $\lambda \in (0,\infty )$ with instantaneous center $C$, and the angle subtended by the arc, $\phi \in (0,\pi )$. Using arc geometry and applying the inextensible constraint, $L=\lambda\ \phi$, we can relate $q$ and curve parameters as

\begin{align}
	L+l_{1} & =\left(\lambda-r\right)\phi\Rightarrow l_{2}=-r\phi\label{eq:l1}\\
	L+l_{2} & =\left(\lambda+\frac{r}{2}\right)\phi\Rightarrow l_{2}=\frac{\phi r}{2}\label{eq:l2}
\end{align}

\begin{figure}[tb] 
	\centering
	\includegraphics[width=1\linewidth]{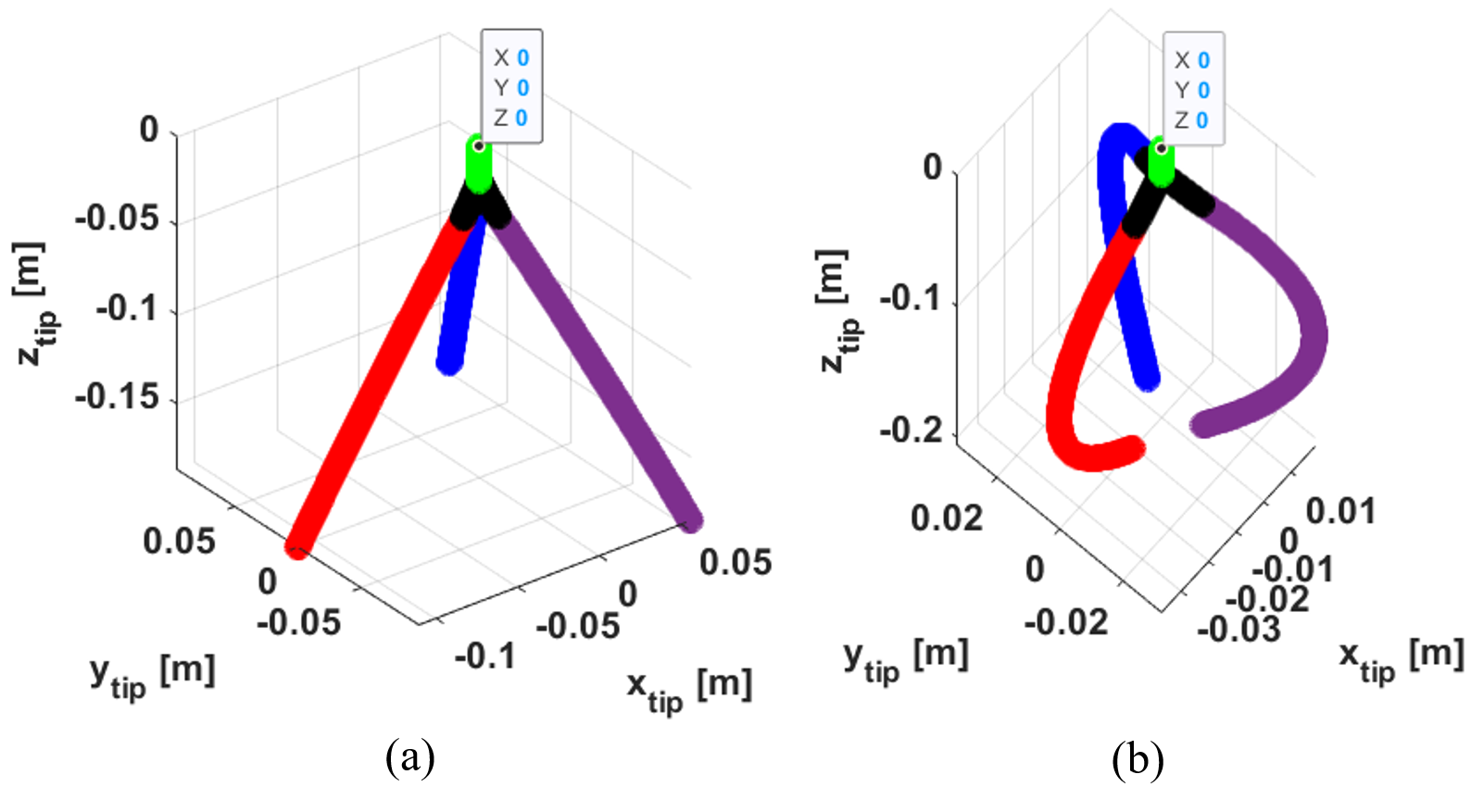}
	\caption{Forward kinematics: (a) Without gripping (b) With gripping.}
	\label{kinematicsimulation} 
\end{figure}

Noting PMAs are extension type and assuming input pressure of PMAs is proportional to their length, we can define $\phi$ as, 
\begin{align}
	\phi & =\begin{cases}
		\frac{-l_{1}}{r} & ;l_{1}>l_{2}\;\text{or}\;P_{1}>P_{2}\\
		\frac{2l_{2}}{r} & ;l_{1}\le l_{2}\;\text{or}\;P_{1}\le P_{2}
	\end{cases}\label{eq:phi}
\end{align}
where $P_i$ is the pressure input to the $i^{th}$ PMA and $\lambda=\frac{L}{\phi}$. Note that, gripping action only involves the scenario $P_1\leq P_2$. Employing the curve parameters, the homogeneous transformation matrix (HTM) can be derived as \cite{godage2015modal}, 

\begin{align}
	\mathbf{T}(c,\xi )=\mathbf{P}_{x}(\lambda)\ \mathbf{R}_{y}(\xi\phi)\ \mathbf{P}_{x}(-\lambda)\nonumber\\
	=\begin{bmatrix} \mathbf{R}(q,\xi)
		& \mathbf{p}(q,\xi) \\  \mathbf{0}_{1x3}
		& \mathbf{1}
	\end{bmatrix} 
\end{align}
where $\mathbf{R}_{y}\in SO(2)$ and $\mathbf{P}_{x}\in \mathbb{R}$ are the homogeneous rotation matrices about $Y$ and translation matrix along the $X$ axes. The scalar $\xi$ define any point along the backbone. $\mathbf{R}\in SO(2)$ and $\mathbf{p}=[x,\ z]^{T}$ denote the rotation and position matrices of the robot finger, respectively. Refer to \cite{godage2015modal} for more details on the derivation. The top unit of the gripper is at the origin of the coordinate frame $\left \{ O \right \}$ as shown in Fig. \ref{schematicdiagram}-b. To derive the kinematic model of the entire gripper, the following translations and rotations are applied.

\begin{align*}
	\mathbf{T}_{1} & =\mathbf{P}_{Z}\left(\sigma\right)\mathbf{R}_{Y}\left(3\pi/4\right)\mathbf{T}\left(q,\xi\right)\\
	\mathbf{T}_{2} & =\mathbf{R}_{Z}\left(2\pi/3\right)\mathbf{T}_{1}\;\;\text{and}\;\;\mathbf{T}_{3}=\mathbf{R}_{Z}\left(4\pi/3\right)\mathbf{T}_{1}
\end{align*}
where $\mathbf{T}_{1}$ is the HTM of the $i^{th}$ finger, and $\mathbf{R}_{z}$ is the homogeneous rotation matrix about the $Z$ axis. The forward kinematics can be used to illustrate gripper operation in the taskspace, as shown in Fig. \ref{kinematicsimulation}.

\begin{figure}[tb] 
	\centering
	\includegraphics[width=\linewidth]{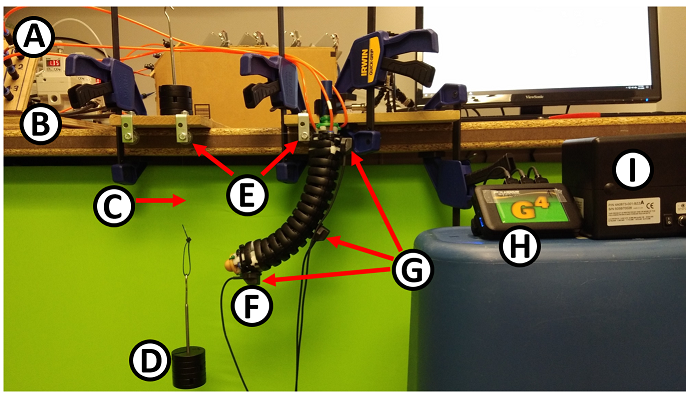}
	\caption{Experiment setup to map bending stiffness, finger shape, and actuation pressure; (A) pressure distribution tubes, (B) proportional digital pressure controllers, (C) Nylon string, (D) load, (E) pulleys, (F) robot finger module, (G) tracking sensor probes, (H) tracking source, and (I) wireless communication hub of the Polhemus G4 wireless magnetic tracking system, respectively.}
	\label{ExperimentSetupStiffnessPressurMapping} 
\end{figure}

\subsection{Map Actuation Pressure, Stiffness, and Shape \label{subsec:mapping}}
We experimented with mapping the finger shape to PMA pressure combinations. We used the experimental setup shown in Fig. \ref{ExperimentSetupStiffnessPressurMapping} to obtain a rich set of data related to shape variation against the actuation pressure combinations. Here, we applied a range of actuation pressures under no-load condition. At each pressure amount, we recorded the bending angle ($\phi$) of the finger using a position and orientation tracking sensor \cite{PolhemusG}. 

\begin{figure}[tb] 
	\centering
	\includegraphics{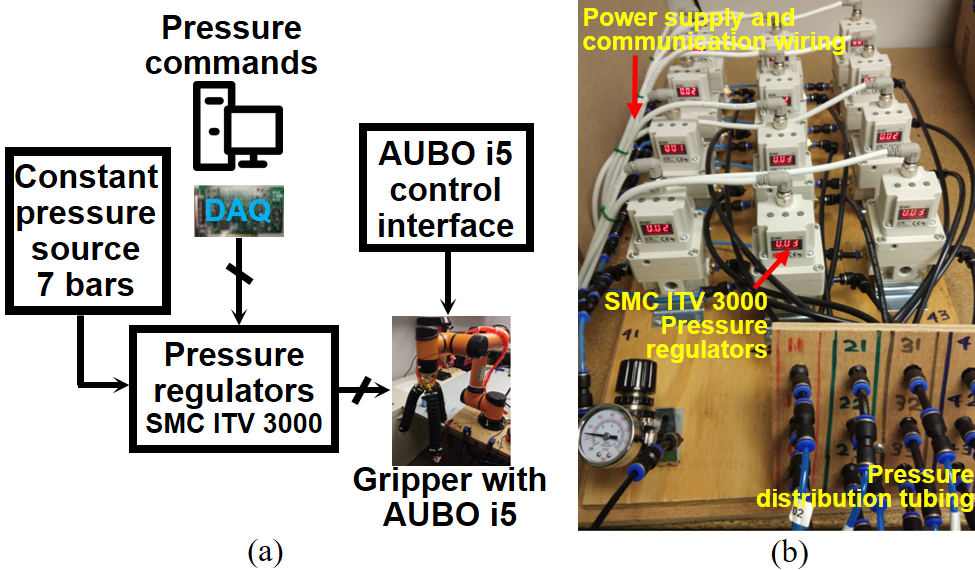}
	\caption{(a) Gripper operation setup, (b) Pressure regulator assembly }
	\label{operationdiagrampressurevalvesetup}
\end{figure}

\begin{figure}[tb] 
	\centering
	\includegraphics[width=1\linewidth]{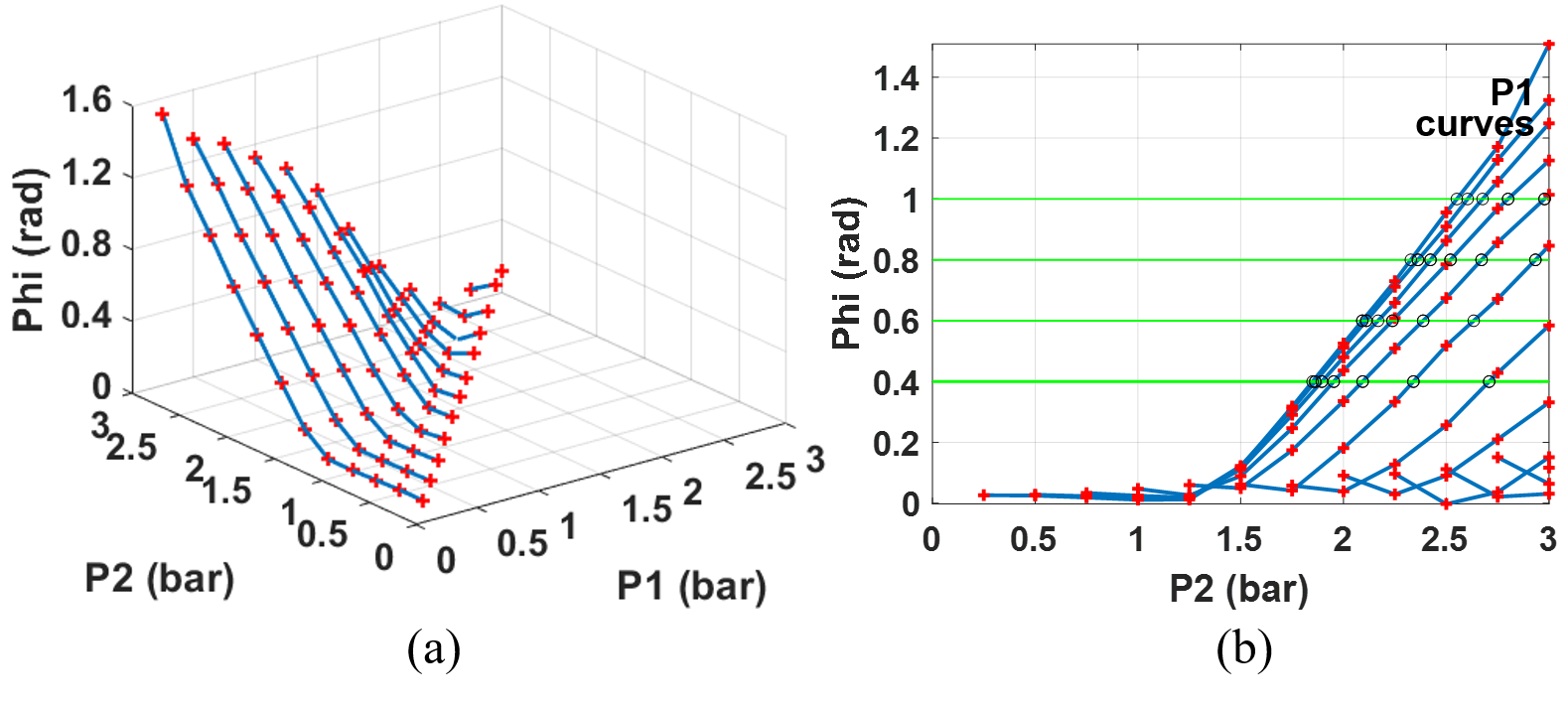}
	\caption{(a) Bending shape variation of a finger, (b) Use the shape-pressure mapping to find four different stiffness values associated with the same bending angle, $\phi$.}
	\label{BendingStiffnessvspressure} 
\end{figure}

Fig. \ref{BendingStiffnessvspressure}-a shows how $\phi$ varies per input pressure combinations. To experiment with the effect of shape and stiffness control, we extract pressure combinations with the same $\phi$. We achieve this by considering planes at different $\phi$ values and interpolating the pressure combinations that result in the desired $\phi$, as shown in Fig. \ref{BendingStiffnessvspressure}-b. The first three columns of Table \ref{table:mappingtable} detail the identified $\phi$ values and the pressure combinations. Next, we used the same experiment set up in Fig. \ref{ExperimentSetupStiffnessPressurMapping} to obtain stiffness values for the identified $\phi$ values and pressure combinations. Here, we applied identified actuation pressures under no-load and fixed-load conditions. At each pressure combination, we recorded the bending angle, change in bending angle ($\Delta \phi$) of the finger, and the torque loading ($\Delta \tau$) that caused the bending angle perturbation. We applied the load profiles using a pulley arrangement so that torque perturbation is normal to the gripper finger module’s neutral axis. Then we calculated the bending stiffness using $K=\frac{\Delta \tau}{\Delta \phi}$ as presented in the final column of Table \ref{table:mappingtable}. The resulting stiffness variation indicates that the bending stiffness increases when the common-mode actuation pressure increases in the PMAs. It is clear that we can obtain the same bending angle for different combinations of $P_1$ and $P_2$ but with different stiffness values indicating the decoupled shape and stiffness control capability. The shape-stiffness mapping will be used to validate the gripper in the next section, experimentally. 

\begin{table}[tb]
	\centering
	\caption{Mapping finger shapes, pressure, and stiffness values}
	\label{table:mappingtable}
	\begin{tabular}{|c|c|c|c|} 
		\hline
		\multirow{2}{*}{\begin{tabular}[c]{@{}c@{}} \textbf{Finger shape}\\{[}\textbf{$\phi$} (rad)]\\ \end{tabular}} & \multicolumn{2}{c|}{\begin{tabular}[c]{@{}c@{}}\textbf{Pressure}\\\textbf{combinations} \end{tabular}} & \multirow{2}{*}{\begin{tabular}[c]{@{}c@{}}\textbf{Bending stiffness}\\(Nm/ rad) \end{tabular}} \\ 
		\cline{2-3}
		& \textbf{P1} (bar)  & \textbf{P2} (bar)  &  \\ 
		\hline
		\multirow{4}{*}{0.4} & 0.50 & 1.86 & 0.63 \\ 
		\cline{2-4}
		& 0.75 & 1.90 & 0.81 \\ 
		\cline{2-4}
		& 1.00 & 1.96 & 1.11 \\ 
		\cline{2-4}
		& 1.25 & 2.09 & 1.32 \\ 
		\hline
		\multirow{4}{*}{0.6} & 0.50 & 2.11 & 0.71 \\ 
		\cline{2-4}
		& 0.75 & 2.17 & 0.85 \\ 
		\cline{2-4}
		& 1.00 & 2.24 & 1.40 \\ 
		\cline{2-4}
		& 1.25 & 2.39 & 1.71 \\ 
		\hline
		\multirow{4}{*}{0.8} & 0.50 & 2.36 & 0.86 \\ 
		\cline{2-4}
		& 0.75 & 2.42 & 1.42 \\ 
		\cline{2-4}
		& 1.00 & 2.52 & 1.90 \\ 
		\cline{2-4}
		& 1.25 & 2.67 & 2.18 \\ 
		\hline
		\multirow{4}{*}{1.0} & 0.50 & 2.60 & 1.56 \\ 
		\cline{2-4}
		& 0.75 & 2.68 & 1.98 \\ 
		\cline{2-4}
		& 1.00 & 2.80 & 2.33 \\ 
		\cline{2-4}
		& 1.25 & 2.98 & 2.58 \\
		\hline
	\end{tabular}
\end{table}

\section{Experimental Validation\label{sec:experimentalvalidation}}

\begin{figure*}[t] 
	\centering
	\includegraphics[width=1\textwidth]{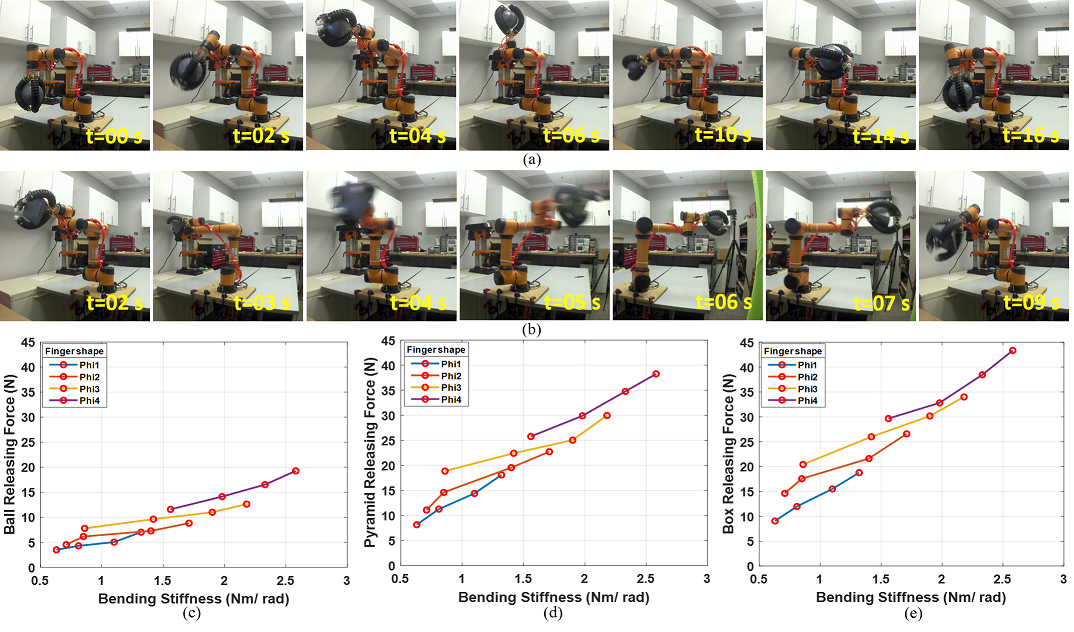}
	\caption{(a) Orientation test sequences of the sphere, (b) Velocity test sequences of the pyramid, (c) sphere, pyramid and box releasing forces}
	\label{orientationvelocityReleasingTests} 
\end{figure*}

\subsection{Experimental Setup\label{subsec:Actuation-Setup}}
The overall gripper operation is illustrated in Fig. \ref{operationdiagrampressurevalvesetup}-a. The setup's input pressure is a constant 7~bar pressure, supplied by a central pneumatic air compressor. It is distributed to SMC ITV3000 series electro-pneumatic pressure regulators \cite{SMCITV} shown in Fig. \ref{operationdiagrampressurevalvesetup}-b. PMAs of the gripper fingers are connected to two pressure regulators to control the bending (the bundled two-PMAs) and stiffness (the remaining PMA). The pressure commands are generated by a Matlab Simulink Desktop Real-time model and communicated via a National Instrument DAQ card. To validate the gripper, we attach the gripper base to the end of an AUBO-i5 \cite{AUBOManual} industrial robot via a custom-designed unit (Fig. \ref{CADPMAFingerGripper}-B). The AUBO-i5 robot manipulator has 6-DoF with maximum joint speeds up to 29.7~rpm and controlled by programs developed using AUBORPE software \cite{AUBOManual}.

\subsection{Testing Methodology\label{subsec:Testing}}
We define the grip quality by three metrics: 1) holding ability as the gripper change orientation, 2) maintaining the grip as the gripper moves, and 3) the amount of force required to release the object from the grip. We conducted three tests to validate each of these qualities, namely orientation test, velocity test, and force test. We used three different object shapes, sphere, box, and pyramid. We selected object shapes to have smooth surfaces as well as a combination of geometrically varying surfaces. We lined the finger surfaces with sandpaper to improve the grip. We cover the object surfaces with the same material in order to provide a uniform friction coefficient. The object sizes were determined based on the finger dimensions. Each object's largest dimension (box, pyramid edge lengths, and sphere diameter) is approximately similar to the finger length. Each object weighs about 100~g. Table \ref{table:mappingtable} provides details on required pressure combinations to obtain sixteen different stiffness values while keeping the finger shape constant on four different occasions. We conducted each of orientation, velocity, and force tests under sixteen shape-stiffness variations. The experiments are repeated for all the objects.

\subsection{Orientation Test\label{subsec:Orientation-Test}}
The orientation test was planned to present how the gripper can handle objects with different profiles and shapes against the gravity and rotational motions. In practical operations, robots have to reorient payloads. Hence a gripper should be able to reorient objects. In this experiment, we move a grasped object from one orientation to another using the two end-effector joints of the manipulator robot running at 15~rpm. A series of images related to the orientation test on a spherical object is shown in Fig. \ref{orientationvelocityReleasingTests}-a. For all objects, the gripper showed firm grasping without failure under all stiffness variations. Therefore, the successful handing of objects with no quantitative difference shows that our gripper design is robust in orientation tasks.

\subsection{Velocity Test\label{subsubsec:Velocity-Test}}
A gripper's ability to handle linear velocity dictates how fast the robot can complete the handling job, i.e., pick and place operation. Thus, to improve productivity, the gripper should be able to handle payloads at higher velocities. We conducted the velocity test to ensure grip security. Here, the robot arm is used to grasp an object and move from one place to another at a higher linear velocity while maintaining the gripper orientation. The velocity test steps of the pyramid are presented in Fig. \ref{orientationvelocityReleasingTests}-b. First, the pyramid is grasped at a pre-programmed home position of the AUBO-i5 robot. Next, a linear trajectory is executed. Similar to the orientation test, we conducted the velocity test for all three objects. The gripper showed firm object holding under all stiffness variations at the maximum moving velocity. Therefore, the successful handling of objects with no quantitative difference shows that the proposed gripper is robust in velocity handling. Our media submission files present the results and both orientation and velocity tests under sixteen stiffness variations.

\begin{figure}[tb] 
	\centering
	\includegraphics[width=1\linewidth]{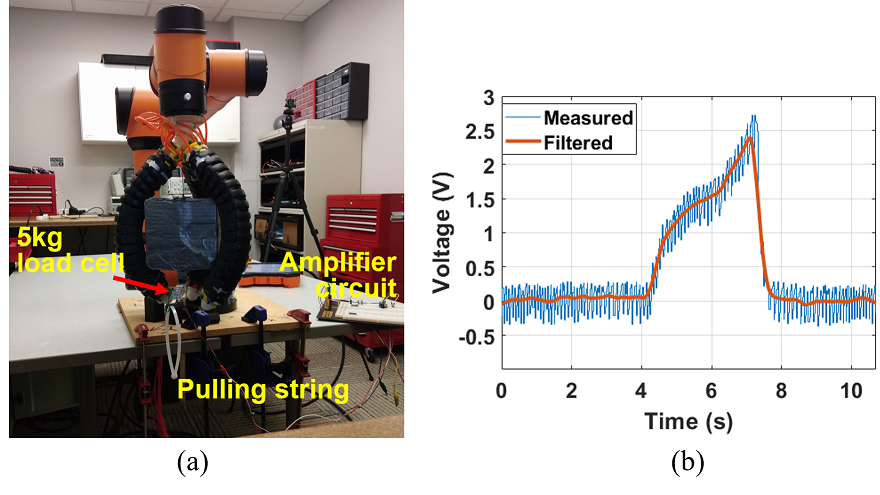}
	\caption{(a) Force test experiment, (b) Maximum force measurement}
	\label{ForceTestmaximumvoltage} 
\end{figure}

\subsection{External Force Perturbation Test\label{subsec:Force-Test}}
The force test is designed to measure the effect of stiffness change quantitatively. In this test, a pulling force is applied until the object released from the grip. Fig. \ref{ForceTestmaximumvoltage}-a shows how the force test is conducted on the box-shaped object. We used a  5~kg load cell coupled with an instrumentation amplifier to measure the pulling force exerted on the object. First, we calibrated the load cell to read standard force values based on voltage readings. The load cell was firmly attached to the object so that a vertical downward force can be applied until the grip fails. Fig. \ref{ForceTestmaximumvoltage}-b illustrates a typical reading recorded during a pulling action. We used a moving average filter to filter noisy data and obtain the force when the grip fails. Similar to the previous two tests, we conducted the force test for all three objects under four finger shapes. Altogether, the releasing force was measured under sixteen stiffness variations. Measured maximum forces against the stiffness variation for each object are presented in Fig. \ref{orientationvelocityReleasingTests}-c. The data shows that, for the same finger shape ($\phi$), high stiffness grasp resulted in a high grip failure force. When the bending angle is high (firmer grip), the pulling force records a higher value. The sphere shows a relatively smoother surface than the pyramid and box surfaces. Therefore, the sphere's failure force at each finger shape shows a relatively lower value than that of the pyramid and the box. The box has the largest geometrically irregular shape, which resulted in the highest failure forces. These extensive results show that the proposed soft gripper design with independent shape and stiffness control can complete various grasping tasks under challenging conditions.

\section{Conclusion}
Due to inherent compliance and conforming capability, soft robotic grippers present a promising path toward adaptive grasping. We hypothesized that independent shape and stiffness control and improve the grasping operation and grip quality. This paper proposed a novel soft robotic gripper based on an inextensible finger design with independent shape and stiffness control. We detailed the proposed soft finger design and how the use of an articulable backbone improves structural integrity and facilitates independent shape and stiffness control. The kinematic models of soft finger modules and the gripper was derived and validated. For better accuracy, we utilized empirical data gathered from rigorous testing to control shape and stiffness. We conducted three tests to show the effectiveness of the proposed soft robotic gripper. The tests demonstrated the gripper's ability to sustain rotational and linear motion with grasped objects. The force test showed that the stiffness control could be used to improve the gripping quality.

\bibliographystyle{IEEEtran}
\bibliography{refs}

\end{document}